%% file: colm2026_conference.tex
\documentclass{article}
\usepackage[preprint]{colm2026_conference}

\usepackage{microtype}
\usepackage{hyperref}
\usepackage{booktabs}
\usepackage{amsmath}
\usepackage{xurl}
\usepackage{graphicx}
\usepackage{colortbl}
\usepackage{booktabs}
\usepackage{multirow}
\usepackage{enumitem}
\usepackage{wrapfig}
\usepackage{pifont}
\usepackage{etoolbox}
\usepackage[normalem]{ulem}
\usepackage[capitalize,noabbrev]{cleveref}

\newcommand{\method}[0]{Cog-DRIFT}

\definecolor{color5}{HTML}{006795}
\hypersetup{
  colorlinks   = true,
  urlcolor     = blue, 
  linkcolor    = color5, 
  citecolor   = color5 
}
\usepackage[framemethod=TikZ]{mdframed}

\definecolor{ExampleBg}{HTML}{ffffff}
\definecolor{ExampleTitle}{HTML}{629677}
\newcounter{example}

\mdfdefinestyle{examplestyle}{
    roundcorner=5pt,
    backgroundcolor=ExampleBg,
    linecolor=ExampleTitle,
    outerlinewidth=0.3pt,
    frametitlebackgroundcolor=ExampleTitle,
    frametitlefont={\bfseries\color{white}},
    skipabove=10pt,
    frametitleaboveskip=3pt,
    frametitlebelowskip=3pt,
}

\newenvironment{example}[1][]{%
    \refstepcounter{example}%
    \ifstrempty{#1}%
        {\def\exampletitle{Example~\theexample}}%
        {\def\exampletitle{Example~\theexample: #1}}%
    \mdframed[style=examplestyle, frametitle=\exampletitle]%
}{%
    \endmdframed
}

\newenvironment{example*}[1][]{%
    \ifstrempty{#1}%
        {\mdframed[style=examplestyle]}%
        {\mdframed[style=examplestyle, frametitle={#1}]}%
}{%
    \endmdframed
}
\crefname{example}{Example}{Examples}

\definecolor{TakeAwayBg}{HTML}{ffffff}
\definecolor{TakeAwayTitle}{HTML}{364156}
\newcounter{takeaway}
\mdfdefinestyle{takeawaystyle}{
    roundcorner=5pt,
    backgroundcolor=TakeAwayBg,
    linecolor=TakeAwayTitle,
    outerlinewidth=0.3pt,
    frametitlebackgroundcolor=TakeAwayTitle,
    frametitlefont={\bfseries\color{white}},
    skipabove=10pt,
    frametitleaboveskip=3pt,
    frametitlebelowskip=3pt,
}

\newenvironment{takeaway}[1][]{%
    \refstepcounter{takeaway}%
    \ifstrempty{#1}%
        {\def\takeawaytitle{Takeaway~\thetakeaway}}%
        {\def\takeawaytitle{Takeaway~\thetakeaway: #1}}%
    \mdframed[style=takeawaystyle, frametitle=\takeawaytitle]%
}{%
    \endmdframed
}
\crefname{takeaway}{Takeaway}{Takeaways}

\usepackage{lineno}

\definecolor{darkblue}{rgb}{0, 0, 0.5}
\hypersetup{colorlinks=true, citecolor=darkblue, linkcolor=darkblue, urlcolor=darkblue}

\title{\centering Cog-DRIFT: Exploration on Adaptively Reformulated\\Instances Enables Learning from Hard Reasoning Problems}

\author{
  \hspace{24pt}
  \shortstack{\\[4pt]
Justin Chih-Yao Chen \quad Archiki Prasad \quad Zaid Khan \quad Joykirat Singh \\[2pt] Runchu Tian \quad Elias Stengel-Eskin \quad Mohit Bansal
    \\[4pt]
    \normalfont
    UNC Chapel Hill
  }
}

\begin{document}

\ifcolmsubmission
\linenumbers
\fi

\maketitle

\begin{abstract}
Reinforcement learning from verifiable rewards (RLVR) has improved the reasoning abilities of large language models (LLMs), yet a fundamental limitation remains: models cannot learn from problems that are too difficult to solve under their current policy, as these yield no meaningful reward signal. 
This creates a performance ceiling where further training yields diminishing returns. 
We propose a simple yet effective solution based on task reformulation. We transform challenging open-ended problems into cognitively simpler variants -- such as multiple-choice and cloze formats -- that preserve the original answer while reducing the effective search space and providing denser learning signals. These reformulations span a spectrum from discriminative to generative tasks, which we exploit to bootstrap learning: models first learn from structured, easier formats, and this knowledge transfers back to improve performance on the original open-ended problems.
Building on this insight, we introduce \method{}, a framework that automatically constructs reformulated variants and organizes them into an adaptive curriculum based on difficulty. 
Training progresses from easier to harder formats as the model improves, enabling the model to learn from problems that previously yielded zero signal under standard RL post-training. \method{} not only improves on the originally unsolvable hard problems (+10.11\% for Qwen and +8.64\% for Llama in absolute gains), but also generalizes well to other held-out datasets. Across two models and six challenging reasoning benchmarks, our method consistently outperforms standard GRPO and strong guided-exploration baselines. On average, \method{} shows +4.72\% (Qwen) and +3.23\% (Llama) improvements over the second-best baseline.
We further show that \method{} improves pass@k at test time, and the curriculum improves sample efficiency.
Overall, our results highlight task reformulation and curriculum learning as an effective paradigm for overcoming the exploration barrier in LLM post-training.\footnote{Code is available at: \url{https://github.com/dinobby/Cog-DRIFT}}
\end{abstract}

\section{Introduction}
Learning from problems that exceed a learner's current capacity is inherently difficult.
This challenge is not unique to models: in human learning, the \emph{Zone of Proximal Development} \citep{vygotsky1978development} describes how learning is maximized when tasks are neither too easy nor too difficult, but instead lie just beyond the learner's current capabilities and can be solved with appropriate scaffolding \citep{wood1976scaffolding,berk1995scaffolding}.
Educational psychology further shows that problem format plays a central role in shaping cognitive load and learning dynamics. Structured formats such as multiple-choice (MCQs) \citep{kelly1916mcq,mcqstable2014} and fill-in-the-blank (cloze) tasks \citep{taylor1953cloze,polat2020analysis} have been used to reduce cognitive load and scaffold learning \citep{clt1988,mcqstable2014, workedexp2004}.
Theoretically, constraining the effective search space and transforming generation into recognition reduces cognitive load \citep{ko2010comparison,howard2017method}. Over time, repeated exposure to such scaffolded tasks supports learning and facilitates transfer to more complex, open-ended problems \citep{workedexp2004,sweller1998design}. Taken together, these insights from human learning suggest a compelling analogy for training large language models (LLMs): when a problem is too difficult to learn from directly, reformulating it into structured variants might bring it into the model's Zone of Proximal Development, enabling learning signals that would otherwise be inaccessible. 

\begin{figure*}[t!]
    \centering
    \includegraphics[width=\linewidth]{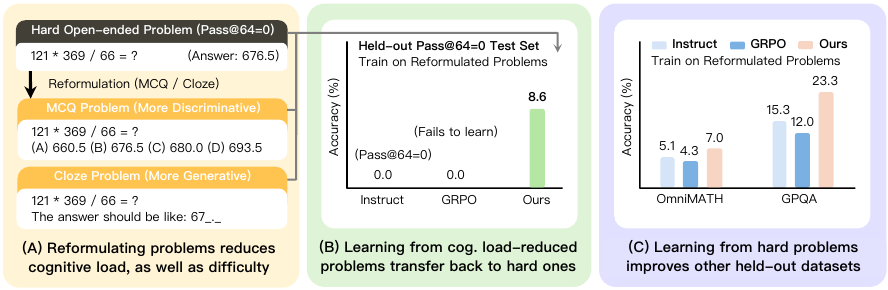}
    \caption{(A) If a problem is too hard (e.g., pass@64=0), the model cannot learn from it. Reformulation into an MCQ or a cloze task can effectively reduce cognitive load, as well as difficulty for the model. (B) We find that learning from cognitive-load-reduced tasks transfers back to the original hard questions (i.e., learning from MCQ/cloze also improves open-ended questions). (C) These improvements on the original hard questions also generalize to held-out datasets. Results in (B) and (C) are based on Llama3.2-3B-Instruct.
    }
    \label{fig:fig1}
\end{figure*}

This perspective is particularly relevant for Reinforcement Learning with Verifiable Reward (RLVR). 
While RLVR has become a dominant learning paradigm to improve LLM reasoning \citep{guo2025deepseek}, a key limitation remains in commonly used algorithms such as PPO \citep{schulman2017proximal} and GRPO \citep{shao2024deepseekmath}, which we visualize in \cref{fig:fig1}(A): when models are trained directly on difficult problems, on-policy RL often fails to obtain any positive reward signal, leading to zero advantages and no meaningful updates. Here, we define ``difficult'' as pass@64=0, following past work \citep{jiang2026fateformalbenchmarkseries}. This failure mode resembles the ``out of the Zone of Proximal Development'' setting in human learning, where tasks are too difficult for learning to occur.
Prior work has attempted to address this issue in two main ways: either by leveraging stronger models to generate expert trajectories for guidance~\citep{qu2025pope}, or by using the same policy to generate either easier problems or hints to reduce problem difficulty \citep{sundaram2026teaching,chen2025nudging,li2025questa,wang2025dissmath}.
However, stronger models are not always available, and soliciting expert guidance can be costly, limiting its scalability. Moreover, hard problems often represent the ceiling of capacity, so obtaining solutions from an expert model or human may not be feasible (i.e., even too hard for experts).
Self-generated hints can be effective, but are also limited by the model's overall capabilities, leaving room for improvement (c.f. \cref{sec:related_work}).

We instead propose task reformulation as a practical solution. 
By rewriting open-ended questions (OEQ) as multiple-choice (MCQ) or fill-in-the-blank (cloze) tasks, the output space is dramatically constrained and learning signals become far denser. 
Importantly, these reformulations span a spectrum between discriminative and generative learning tasks. MCQs are more discriminative, turning the task into one of ranking or selection. Cloze tasks are partially generative, requiring the model to produce missing components of the answer. OEQs lie at the most generative extreme.
Our approach leverages this spectrum by using discriminative tasks to bootstrap learning: generative tasks teach models to avoid shortcuts (e.g., skipping reasoning steps, outputting the correct answer choice with partial input) \citep{balepur2025these,chandak2025answer,li2025generative} while discriminative tasks are easier to learn \citep{geirhos2020shortcut}.
In \cref{fig:fig1}(B) and \cref{sec:preliminary}, we find that once the model learns on the reformulated tasks, it can transfer back to the original open-ended setting.
However, we find that naively using a single reformulation is insufficient for generalization.
Empirically, performance improvements on held-out datasets only emerge when models are exposed to multiple reformulations simultaneously, suggesting that diversity in difficulty is key.
This observation motivates \method{}, a framework that reformulates hard problems into a spectrum of variable difficulty problems, and organizes them into an adaptive curriculum analogous to cognitive scaffolding approaches in human education.\footnote{We use the term ``cognitive'' to draw an analogy to human learning, referring specifically to structured progression in task difficulty rather than implying human-like cognition in models.}
Concretely, given a challenging open-ended question, we generate a family of cognitively simplified variants (e.g., multiple-choice with varying numbers of options, cloze-style questions), all of which preserve the original answer and thus maintain verifiable correctness. Rather than viewing these reformulations as isolated augmentations, we interpret them as forming a difficulty spectrum to enable ordered progression. This design also aligns with the Zone of Proximal Development in human learning, where learners benefit most from tasks that are appropriately difficult.
\method{} leverages this structure by adaptively advancing the model from easier to harder variants based on empirical accuracy (i.e., progressing to the next level once a target accuracy threshold is reached), thereby automating curriculum construction. We show that compared to static mixtures of reformulations, this adaptive curriculum yields more sample-efficient training.

We evaluate \method{} on six benchmarks, including BigMath-Hard \citep{albalak2025big}, OmniMATH-Hard \citep{gao2024omni}, AIME2024/2025 \citep{AIME}, GPQA-Diamond \citep{rein2024gpqa} and Date Understanding \citep{srivastava2022beyond} with two models: Qwen3-4B-Instruct-2507 \citep{qwen3} and Llama3.2-3B-Instruct \citep{llama3}. 
Results demonstrate that reformulating hard problems to reduce cognitive load successfully unlocks learning where standard methods fail: we find that directly training on the hard problems using GRPO is not stable with sparse reward signals, and can harm generalization on held-out tasks. 
In contrast, \method{} yields consistent gains across benchmarks: compared to the second-best baseline, our method shows an absolute +4.72\% average improvement for Qwen, and +3.23\% for Llama. 
Moreover, \method{} also improves pass@k over both the base model and the GRPO-trained variant, suggesting that it enables the acquisition of new reasoning capabilities from hard examples that initially lie beyond the model's capacity.
We conduct extensive analyses and highlight two critical factors driving this success: (1) exposing the model to a diverse mixture of reformulations (such as combining multiple-choice and cloze formats) is essential for transferring reasoning skills back to open-ended tasks, and (2) an instance-level curriculum that progressively scales problem difficulty significantly boosts sample efficiency and prevents performance plateaus.

\section{Related Work}
\label{sec:related_work}
\textbf{Reinforcement Learning with Verifiable Rewards (RLVR).} RLVR has emerged as a dominant paradigm for training large language models on reasoning tasks by utilizing objective, rule-based feedback rather than human preferences \citep{lambert2024tulu, guo2025deepseek, team2025kimi, zeng2025simplerl}. By employing deterministic verifiers (e.g., binary rewards based on final answer correctness), RLVR is particularly effective for mathematics and code generation where ground truth is unambiguous. 
Extending this paradigm to unverifiable domains, \cite{lu2026golden} demonstrate that large-scale internet text can be transformed into structured formats such as multiple-choice and masked prediction tasks to enable RLVR.
Central to the current discourse is whether RLVR enables the genuine discovery of new reasoning abilities or primarily performs "distribution sharpening" by amplifying high-reward paths already present in the model's latent space \citep{zhang2025rlsurvey, zhao2025echo, shenfeld2025rl}. While some studies suggest RLVR mainly improves sampling efficiency (pass@1) without increasing the total knowledge (pass@$k$) \citep{yue2025does, he2025rewarding}, others argue that RL can indeed foster new skills through the strategic combination of existing ones \citep{yuan2025llms} or by balancing the entropy-reward trade-off to prevent premature convergence \citep{wu2025invisible}. Although with a sufficiently large $k$ (e.g., $\text{vocab\_size} ^ \text{seq\_len}$), even a randomly initialized model could achieve near-perfect pass@$k$, RLVR operates under a rollout budget, so how to improve the model within a limited sampling budget, where $k$ must remain small relative to the total number of rollouts becomes the real question \citep{yu2026unveiling,nath2025adaptive}.

\noindent\textbf{Learning from the Hard Problems.} 
A challenge in RLVR occurs when problems are excessively difficult such that on-policy sampling fails to discover any correct solutions, i.e., the model cannot get any non-zero reward \citep{qu2025pope}. To bridge this gap, research has shifted toward sophisticated exploration strategies and hybrid optimization. One line of work focuses on "nudging" the model through self-generated hints, critiques, or partial solutions to reduce problem difficulty and expand the reasoning boundary \citep{li2025questa, zhang2025critique, chen2025nudging, wang2025dissmath}. 
\cite{ma2025learning} interleave SFT to learn from hard questions that RLVR cannot.
\cite{sundaram2026teaching} train the teacher policy for generating question-answer pairs based on the improvement of the student policy.
Other approaches leverage privileged information or offline trajectories to provide a learning signal in sparse-reward environments \citep{lu2025arpo, yan2025learning}, such as POPE, which uses offline prefixes from a stronger model to guide exploration into reachable reward regions \citep{qu2025pope}. 
In parallel, \cite{sun2025rl} propose a ``grokking'' recipe for RLVR by showing that staged warm-ups, experience replay, and a curriculum enable learnability on previously unsolvable problems.
Together, these advancements suggest that scaling RLVR requires moving beyond pure on-policy exploration toward self-improvement mechanisms and careful curriculum design \citep{shi2025efficient,parashar2025curriculum,bae2026online}.
Despite their differences, these approaches share a common principle: they aim to make hard problems learnable by injecting additional signal -- either through stronger supervision, guided exploration, or richer training dynamics.
Our approach takes a fundamentally different perspective. We introduce cognitive load as a unifying lens and show that reformulating problems into simpler variants enables the model to access reward signals that are otherwise unreachable under standard RLVR. Crucially, this is coupled with an adaptive curriculum that dynamically schedules these reformulations based on the model’s learning progress, ensuring that training remains within a learnable regime while gradually increasing difficulty. Empirically, this leads to a substantial increase in the rate of hard problems during training that become solvable, reaching up to 18.9\%, compared to the 1-4\% range reported by prior methods \citep{chen2025nudging, sundaram2026teaching}.

\section{Problem Statement}
\label{sec:preliminary}
Let $\mathcal{M}_{\theta}$ denote a language model with parameters $\theta$. We consider a dataset of open-ended problems $\mathcal{D} = \{(x_i, y_i)\}_{i=1}^{N}$, where $x_i$ is a problem prompt and $y_i$ is the corresponding exact answer.\footnote{We use ``open-ended'' to denote problems with unconstrained output length, in contrast to multiple-choice or short-form tasks, while still assuming a well-defined correct answer.} The model can generate $k$ independent samples $\{\hat{y}_i^{(j)}\}_{j=1}^{k}$ for a specific $x_i$, and the commonly used pass@$k$ metric evaluates whether at least one of the $k$ generations is correct: 

$$\text{pass@}k(x_i) = \mathbb{I}[\exists j \in \{1,\dots,k\} \text{ such that } \hat{y}_i^{(j)} = y_i ].$$

We define the set of \emph{hard problems} as the subset of examples on which the model fails to produce a correct answer even after 64 samples: $\mathcal{D}_{\text{hard}} = \{(x_i, y_i) \in \mathcal{D} \mid \text{pass@64}(x_i) = 0\}.$
This definition has an important implication for RL training with rollout-based objectives (e.g., GRPO): if $\text{pass@64}(x_i)=0$, then the probability of sampling a correct trajectory is extremely low (requiring $\gg 64$ samples) given the same policy. 
Consequently, the reward signal is generally non-existent: 
all sampled trajectories receive zero reward, 
leading to zero variance in the advantage estimates. 
This in turn prevents meaningful gradient updates, as no trajectory is preferred over others \citep{qu2025pope, yu2025dapo}. 
Empirically, we show in \cref{sec:analysis} that directly applying GRPO on $\mathcal{D}_{\text{hard}}$ yields little to no improvement because the model rarely explores the correct trajectory in an unconstrained, open-ended output space.
Moreover, training on $\mathcal{D}_{\text{hard}}$ wastes computational effort performing rollouts on solutions with no hope of being correct.
To address this, we introduce a task reformulation function $\mathcal{T} : (x, y) \rightarrow (x', y)$ that converts an open-ended problem into a cognitively simplified format while preserving the underlying solution. A successful reformulation should meet three criteria shown in \cref{tk:criteria}: 

\begin{takeaway}[Three Criteria for an Ideal Reformulation]
\label{tk:criteria}
\begin{enumerate}[leftmargin=*, itemsep=1pt, topsep=0pt]
    \item \textbf{Answer preservation:} must have the same correct answer as the original.
    \item \textbf{Reduced difficulty:} should make the problem empirically easier for the model.
    \item \textbf{Transferability:} Learning from the reformulated problem should improve performance on the original problem.
\end{enumerate}
\end{takeaway}

We explore several instantiations of $\mathcal{T}$, utilizing an LLM (Qwen3-4B-Instruct-2507) to generate the reformulated dataset $\mathcal{D}'_\text{hard} = \{(x'_i, y_i) \mid (x_i, y_i) \in \mathcal{D}_{\text{hard}}\}$. We provide the prompt for reformulation in \cref{sec:prompt}. The formats we consider include:
(1) \textbf{4-Choice:} The original question is combined with four answer options, where the gold answer appears in exactly one option.
(2) \textbf{10-Choice:} Same as above, but we expand the number of option choices to ten.
(3) \textbf{Cloze:} The original question is combined with a partially masked answer. For example, if the answer is 101, we append ``The answer should look like: 1\_\_''. \cref{exp:oeq_4mcq} is an example of an open-ended problem reformulated into a 4-choice format, and other examples for 10-choice and cloze can be found in \cref{sec:problem_type}. Crucially, even after reformulation, the model must output the full answer value (e.g., 21) rather than just the option letter (e.g., B) for better transferability. 

\begin{example}[An Example of $\mathcal{T}$ (Open-ended to Four Choice)]
\label{exp:oeq_4mcq}
\textbf{Question (x):} What is the largest number of cells in a $9 \times 9$ board that can be cut along both diagonals without the board falling into pieces? \textbf{Answer (y): 21}

\textbf{Reformulated Question (x'):} What is the largest number of cells in a $9 \times 9$ board that can be cut along both diagonals without the board falling into pieces?\\A. 18 B. 21 C. 24 D. 15 \textbf{Answer (y): 21}
\end{example}

\begin{figure*}
\centering
\includegraphics[width=\linewidth]{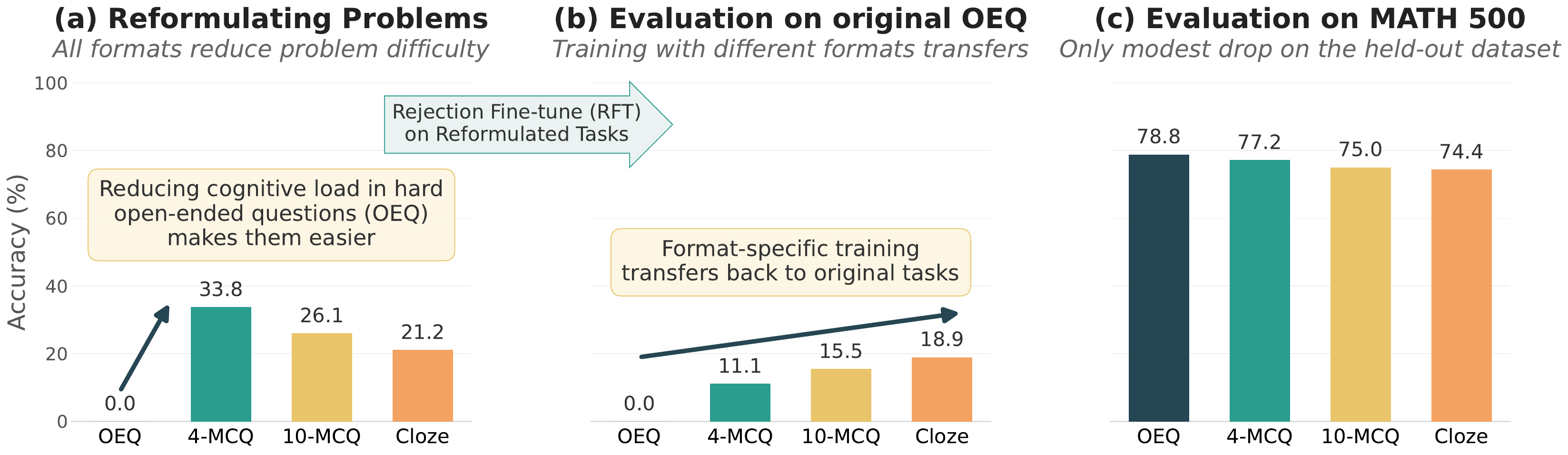}
\caption{(a) Reformulating open-ended math problems into alternative formats consistently increases accuracy by easing structural constraints. (b) After Rejection Fine-Tuning (RFT), these performance gains successfully transfer back to the original open-ended problems used for training. (c) On the unseen MATH 500 benchmark, the model trained with an easier format incurs only a modest drop.}
\label{fig:preliminary}
\end{figure*}

\subsection{Evaluating Difficulty Reduction}
By definition, all of our chosen reformulations satisfy criterion (1) by sharing the original gold answer in \cref{tk:criteria}. To evaluate criterion (2), we measure zero-shot performance on the reformulated dataset $\mathcal{D}'_\text{hard}$, as shown in \cref{fig:preliminary}(a). Because we exclusively sampled problems where pass@64 = 0, the baseline performance on the original open-ended questions is strictly 0.0. However, upon applying format changes, performance immediately surpasses the random guessing baseline. For instance, the 4-choice MCQ reformulation achieves 33.8\% accuracy, well above the 25\% random baseline. Similar trends hold for the 10-choice and cloze formats. We also observe that task reformulations induce a consistent ordering of empirical difficulty: $\text{MCQ}_{4} \;\leq\; \text{MCQ}_{10} \;\leq\; \text{Cloze} \;\leq\; \text{Open-ended}.$

\subsection{Transferability and Generalization}
Having established that these formats reduce difficulty, we evaluate criterion (3) in \cref{tk:criteria}: whether a model can learn from the converted problems and transfer those gains back to the original open-ended ones. To test this, we use $\mathcal{D}'_\text{hard}$ for rejection fine-tuning (RFT) \citep{yuan2023scaling}, as depicted in \cref{fig:preliminary}(b). Specifically, we use $\mathcal{D}'_\text{hard}$ to elicit correct chain-of-thought \citep{wei2022chain} reasoning trajectories, perform supervised fine-tuning (SFT) on the correct $\{x', y\}$ pairs, and then evaluate the trained model on the original open-ended dataset $\mathcal{D}_{\text{hard}}$.
While acquiring correct reasoning chains directly from $\mathcal{D}_{\text{hard}}$ is not possible, the reduced difficulty of $\mathcal{D}'_\text{hard}$ allows us to successfully extract valid trajectories. The results show that models trained on $\mathcal{D}'_\text{hard}$ successfully transfer their learning back to $\mathcal{D}_{\text{hard}}$. 
Interestingly, in \cref{fig:preliminary}(b), we also observe that the more the task is generative, the better the transfer back to the open-ended problem. For example, on the original $\mathcal{D}_{\text{hard}}$, model trained with 4-choice achieves 11.1\%, 10 choice achieves 15.5\%, and cloze achieves 18.9\% accuracy.
To isolate genuine reasoning capabilities from memorization, we evaluate the format-specific trained models on the unseen MATH500 benchmark \citep{lightman2023let,hendrycksmath2021}. As shown in \cref{fig:preliminary}(c), models trained with different reformulations maintain fairly robust performance on this held-out dataset. 
Overall, the MCQ and cloze formats meet the criteria in \cref{tk:criteria}, and successfully (1) preserve the original answer, (2) reduce empirical difficulty, and (3) encourage the model to learn transferable, generalizable reasoning. Therefore, we select these formats for our main experiments.

\section{\uline{Cog}nitive \uline{D}emand \uline{R}eduction via \uline{I}ncremental \uline{F}ormat \uline{T}ransitions}
\label{sec:method}
We build on the observation from \cref{sec:preliminary} that different task reformulations form an order of empirical difficulty: $\text{MCQ}_{4} \;\leq\; \text{MCQ}_{10} \;\leq\; \text{Cloze} \;\leq\; \text{Open-ended}.$
We leverage this structure to design a curriculum that adaptively increases problem difficulty based on the model's competence. Specifically, given a hard problem $(x_i, y_i) \in \mathcal{D}_{\text{hard}}$, we construct a sequence of reformulated variants: $\{x_i^{(1)}, x_i^{(2)}, x_i^{(3)}, x_i^{(4)}\}$, where each level corresponds to a format of increasing difficulty:
\[
x_i^{(1)}: \text{4-choice MCQ}, \quad
x_i^{(2)}: \text{10-choice MCQ}, \quad
x_i^{(3)}: \text{Cloze}, \quad
x_i^{(4)}: \text{Open-ended}.
\]
All variants share the same ground-truth answer $y_i$ and are generated via $\mathcal{T}$ introduced in \cref{sec:preliminary}.
We verify each reformulation deterministically via verification scripts to ensure correctness and validity. 
For MCQ, we verify that: (1) the gold answer appears among the options, and (2) it is uniquely correct (i.e., no other option corresponds to the gold answer). 
For cloze, we ensure that: (1) the masked span is non-trivial (i.e., not equivalent to the full gold answer), and (2) each revealed digit is consistent with the gold answer. 
For example, a gold answer of 101 and mask of 1\_0 would constitute a failure case.

\noindent\textbf{Instance-level Curriculum.}
We associate each training example with a difficulty level $d_i \in \{1,2,3,4\}$, initialized to the easiest level $d_i = 1$. During training, the model $\mathcal{M}_\theta$ generates $m$ samples for the current format:
$\{\hat{y}_i^{(j)}\}_{j=1}^m \sim \mathcal{M}_\theta(\cdot \mid x_i^{(d_i)})$.
We estimate the empirical accuracy at level $d_i$ as:
$\hat{a}_i = \frac{1}{m} \sum_{j=1}^{m} \mathbb{I}[\hat{y}_i^{(j)} = y_i]$.
If $\hat{a}_i \geq \tau$, we promote the example to the next difficulty level.
We adopt $\tau = 0.5$ in our experiments following past work \citep{chen2025self,foster2025learning}.
Otherwise, the example remains at the current level. This creates a per-sample curriculum that adapts to the model's evolving capability, ensuring that training focuses on the most informative difficulty for each problem.

\noindent\textbf{GRPO Training Objective.}
Given recent advancements driven by RLVR, we optimize the model using Group Relative Policy Optimization (GRPO) with rule-based rewards \citep{guo2025deepseek, shao2024deepseekmath}. For each input $x_i^{(d_i)}$, we sample a group of $m$ responses and compute rewards for each sample. Let $r_i^{(j)}$ denote the reward for $\hat{y}_i^{(j)}$. GRPO normalizes rewards within the group, and the objective is: 
$$\mathcal{L}_{\text{GRPO}}(\theta) = - \mathbb{E}_{i,j} \left[ \tilde{r}_i^{(j)} \log \mathcal{M}_\theta(\hat{y}_i^{(j)} \mid x_i^{(d_i)}) \right]$$
which encourages the model to assign higher probability to relatively better samples within each group. 
Our reward function combines correctness and format compliance:
$r_i^{(j)} = r_{\text{correct}}(\hat{y}_i^{(j)}, y_i) + r_{\text{format}}(\hat{y}_i^{(j)})$,
where $r_{\text{correct}} \in [0,1]$ indicates whether the predicted answer matches the ground truth, $r_{\text{format}} \in [0,0.2]$ measures whether the output follows the required format, i.e., whether the final answer is found in \texttt{\textbackslash boxed\{\}} \citep{guo2025deepseek}.

\section{Experimental Setup}
\textbf{Models.}
We evaluate \method{} on two models: Llama3.2-3B-Instruct \citep{llama3} and Qwen3-4B-Instruct-2507 \citep{qwen3}.\footnote{Hereafter, we refer to them as Llama and Qwen, respectively.}
We consider these two models because they show different behaviors in post-training \citep{shao2025spurious}.

\noindent\textbf{Dataset.} 
For training, we use the BigMath dataset \citep{albalak2025big} due to its diversity and inclusion of difficulty annotation. Specifically, it contains 251k samples along with the pass rate using Llama3-8B, making it easier to start with a harder subset with a lower pass rate, and sample 64 reasoning chains to obtain our $\mathcal{D}_\text{hard}$. We start with the hardest 20\% samples (37.5k), and let Qwen sample 64 reasoning chains, only keeping the samples for which \emph{none of the 64 chains have reached the correct solution}. Since Qwen is generally outperforming Llama, we use Qwen to obtain $\mathcal{D}_\text{hard}$ and use the same $\mathcal{D}_\text{hard}$ for Llama. This results in 8.9k samples. We notice that using pass@64 alone causes selection bias to include problems that are actually unsolvable, such as incomplete questions or referring to a figure that does not exist (examples shown in \cref{sec:unsolvable}). To alleviate this, we keep problems for which GPT5.4's majority answer over three runs agrees with the gold answer \citep{gpt54}. This leads to our final filtered training data, consisting of 958 hard problems (more details in \cref{sec:analysis}).
For evaluation, we test \method{} on six benchmarks, including \textbf{(1) BigMathHard (BMH)}, \textbf{(2) OmniMATH-Hard}, \textbf{(3) AIME 24/25}, \textbf{(4) GPQA Diamond}, \textbf{(5) Date Understanding}. Details on each evaluation dataset are provided in \cref{sec:dataset}.

\noindent\textbf{Baselines.}
We mainly compare with baselines that do not need privileged information or require a stronger teacher model:
(1) \textbf{Zero-shot:} We prompt the model to think step-by-step and provide the answer within \textbackslash boxed{\{}\} \citep{kojima2022large}.
(2) \textbf{Few-shot:} Besides prompting the model to think step-by-step, we include 8 in-context learning samples.
(3) \textbf{Rejection sampling Fine-Tuning (RFT):} We prompt the model with cloze reformulated problems, and use the correct trajectories to perform supervised fine-tuning \citep{yuan2023scaling}. We choose fill-in-the-blank here because it performs the best (c.f. \cref{fig:preliminary}).
(4) \textbf{GRPO}: Train on $\mathcal{D_\text{hard}}$ using outcome correctness as the reward \citep{guo2025deepseek}.
(5) \textbf{NuRL (Abstract)}: we use abstract cues that are generated from answer explanations using $\mathcal{D_\text{hard}}$ to hint the model during GRPO \citep{chen2025nudging}.
(6) \textbf{NuRL (Prefix)}: we use partial steps as hints instead, similar to POPE \citep{qu2025pope} except the steps are also self-generated.

\begin{table*}[t]
\centering
\small
\renewcommand{\arraystretch}{1.2}
\begin{tabular}{lccccccc}
\toprule
 & BMH & Omni-Hard & AIME24 & AIME25 & GPQA & Date & Avg. \\
\midrule
\multicolumn{8}{c}{\textbf{Qwen3-4B-Instruct-2507}} \\
\midrule
Zero-shot                    & 0.0  & 15.67 & 46.92 & \textbf{41.71} & 23.57 & 87.71 & 35.93 \\
Few-shot                     & 0.0  & 16.31 & 44.78 & 37.98 & 26.12 & 88.37 & 35.59 \\
RFT                          & 4.32 & 15.13 & 40.06 & 34.88 & 23.04 & 85.11 & 33.76 \\
GRPO                         & 6.37 & \underline{16.37} & 31.87 & 27.29 & \underline{31.81} & \textbf{89.53} & 33.87 \\
NuRL (Abstract)              & \underline{7.73} & 16.03 & \underline{44.81} & 36.44 & 24.72 & 87.48 & \underline{36.20} \\
NuRL (Prefix)                & 6.51 & 15.69 & 45.31 & 38.81 & 22.27 & 84.17 & 35.46 \\\midrule
\rowcolor{blue!7}\method{}                    & \textbf{10.11} & \textbf{18.58} & \textbf{51.74} & \underline{40.28} & \textbf{36.01} & \underline{88.75} & \textbf{40.92} \\
\midrule
\multicolumn{8}{c}{\textbf{Llama3.2-3B-Instruct}} \\
\midrule
Zero-shot                    & 0.0  & 5.09 & \underline{3.75} & 0.42 & 15.32 & 54.87 & 13.24 \\
Few-shot                     & 0.0  & 5.74 & 3.33 & 0.33 & \underline{17.89} & 55.63 & 13.82 \\
RFT                          & \underline{2.47} & 6.52 & 1.67 & 0.00 & 16.13 & 50.31 & 12.85 \\
GRPO                         & 0.0  & 4.32 & 0.68 & 0.42 & 12.03 & 38.85 & 9.92 \\
NuRL (Abstract)              & 1.38 & 6.67 & \textbf{4.17} & \textbf{1.11} & 16.98 & \underline{56.72} & \underline{14.51} \\
NuRL (Prefix)                & 1.04 & \textbf{7.19} & 3.33 & 0.42 & 15.79 & 54.91 & 13.49 \\\midrule
\rowcolor{blue!7}\method{}                    & \textbf{8.64} & \underline{7.04} & \textbf{4.17} & \underline{0.42} & \textbf{23.34} & \textbf{62.82} & \textbf{17.74} \\
\bottomrule
\end{tabular}
\caption{Performance comparison on six benchmarks. We \textbf{bold} the best and \underline{underline} the second-best results. \method{} achieves the strongest overall results, improving avg. accuracy by +4.72\% (Qwen) and +3.23\% (Llama). Crucially, it unlocks gains on BMH (0 → 10.11\% on Qwen / 8.64\% on Llama), where all the problems are collected with pass@64=0.}
\label{tab:main}
\end{table*}

\noindent\textbf{Implementation Details.}
We evaluate all methods using pass@1. Results are averaged over 16 runs for AIME, and 3 runs for other datasets.
We adopt verl \citep{sheng2024hybridflow} as the backbone, and utilize vllm \citep{kwon2023efficient} to speed up rollout generation and inference. To verify equivalence between predictions and references, we use Math-Verify.\footnote{\url{https://github.com/huggingface/Math-Verify}}
Other hyperparameters and details can be found in \cref{sec:implementation} and in our code.

\section{Results and Analysis}
\label{sec:analysis}
\noindent\textbf{Reformulation enables learning from the hard problems.}
In \cref{tab:main}, zero-shot baselines achieve 0\% performance on BigMath-Hard (BMH), as expected for a pass@64=0 dataset. As discussed in \cref{sec:preliminary}, while RFT improves BMH performance, it generalizes poorly due to distribution shift from cloze-style training.
In principle, GRPO should not yield learning gains when trained on samples with pass@64 = 0. However, with Qwen, we observe a +6.37\% improvement on BMH. This may be attributed to the spurious reward effect specific to Qwen \citep{shao2025spurious}, where learning can occur even under weak or uninformative reward signals. 
Another hypothesis is that occasional successful rollouts (arising from stochastic exploration) introduce sparse but amplified reward signals, causing the model to overfit to a narrow subset of problems without learning transferable reasoning strategies.
In contrast, Llama shows no improvement under GRPO, with BMH performance remaining at 0\% after training.
Despite gains on Qwen, GRPO harms generalization, causing drops across held-out benchmarks. 
NuRL with abstract hints (Abstract) or partial solutions (Prefix) provides a middle ground: it yields modest BMH gains and occasional improvements on held-out datasets, likely due to preserving format and leveraging in-domain, self-generated hints.
By contrast, \method{} outperforms all baselines, improving BMH by an absolute +10.11\% (Qwen) and +8.64\% (Llama), while also boosting five held-out benchmarks with average gains of 4.72\% and 3.23\%, respectively. This demonstrates that reformulation enables effective learning with strong cross-task generalization.

\begin{figure*}[t!]
    \centering
    \includegraphics[width=\linewidth]{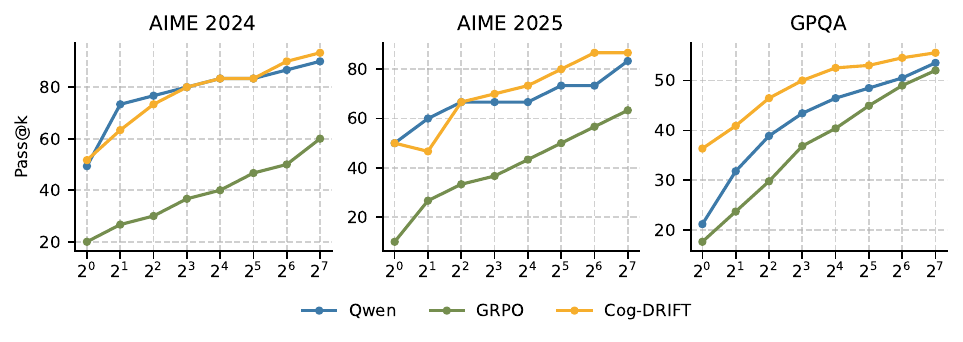}
    \vspace{-22pt}
    \caption{When training on hard open-ended problems and evaluating on AIME24, AIME25, and GPQA using Qwen, \method{} generally achieves higher pass@k compared to both the base Qwen model and the GRPO-trained model, particularly as k increases.}
    \label{fig:pass_at_k}
\end{figure*}

\noindent\textbf{\method{} improves pass@k.}
In \cref{tab:main}, we observe that \method{} consistently outperforms baseline methods in terms of pass@1. Because our approach specifically targets problems with pass@64 = 0 -- i.e., problems that are unsolvable under standard training and therefore contain knowledge inaccessible to the model -- we hypothesize that successful learning from these examples should also manifest as improvements in pass@$k$ for larger values of $k$, reflecting the acquisition of new knowledge.
To test this hypothesis, we plot pass@$k$ for $k \in {1, 2, \ldots, 128}$ in \cref{fig:pass_at_k} using Qwen. On AIME2024, AIME2025, and GPQA, \method{} consistently achieves higher pass@$k$ as $k$ increases, outperforming both the base instruct model and the GRPO-trained model. At $k = 128$, \method{} improves over the base model by +3.33\%, +3.33\%, and +2.02\% on AIME2024, AIME2025, and GPQA, respectively. In contrast, GRPO again exhibits degraded performance, echoing the finding that directly training on hard problems fails to generalize.
These results indicate that learning from reformulated problems reshapes the model's output distribution, making correct solutions more accessible under sampling. Overall, the gains in pass@$k$ provide further evidence that training on reformulated hard problems can effectively learn reasoning capabilities.

\begin{wraptable}{r}{0.48\linewidth}
\small
\centering
\vspace{-10pt}
\begin{tabular}{lcc}
\toprule
\textbf{Data Source} & \textbf{Omni} & \textbf{MATH} \\
\midrule
- & 15.67 & 78.80 \\\midrule
OEQ & 14.16 & 76.20 \\
4-MCQ & 14.82 & 79.80 \\
Cloze & 16.15 & 80.20 \\
\midrule
4-MCQ + Cloze & \textbf{18.36} & \textbf{82.60} \\
\midrule
OEQ + 4-MCQ & 12.39 & 75.40 \\
OEQ + Cloze & 12.61 & 79.80 \\
OEQ + 4-MCQ + Cloze & \textbf{18.42} & \textbf{82.20} \\
\bottomrule
\end{tabular}
\caption{\textbf{Combining multiple reformulations} (e.g., 4 Choice + Cloze, or Original + 4 Choice + Cloze) leads to the largest gains, suggesting that \textbf{diversity of reformulations is critical for learning from hard problems}.}
\label{tab:data_mixture}
\end{wraptable}

\noindent\textbf{Using multiple reformulations is crucial.}
To better understand the transfer effect from reformulated variants to the original open-ended problems, we analyze how different data mixtures influence performance. Specifically, we remove the curriculum setup and instead rejection fine-tune (RFT) the models on various combinations of problem types.
\cref{tab:data_mixture} reports the performance on OmniMATH-Hard and MATH500 under different mixtures. The first three rows show that training with a single reformulation does not improve over no training. Notably, training solely on \textit{4 Choice + Cloze} without any open-ended data achieves improved performance, and it is comparable to mixtures that include open-ended problems.
Therefore, we conclude that diversity in problem representation is critical for effective transfer and for learning from hard problems, and that including more discriminative formats (e.g., \textit{4 Choice + Cloze}) is crucial.

\begin{figure*}[ht!]
    \centering
    \includegraphics[width=\linewidth]{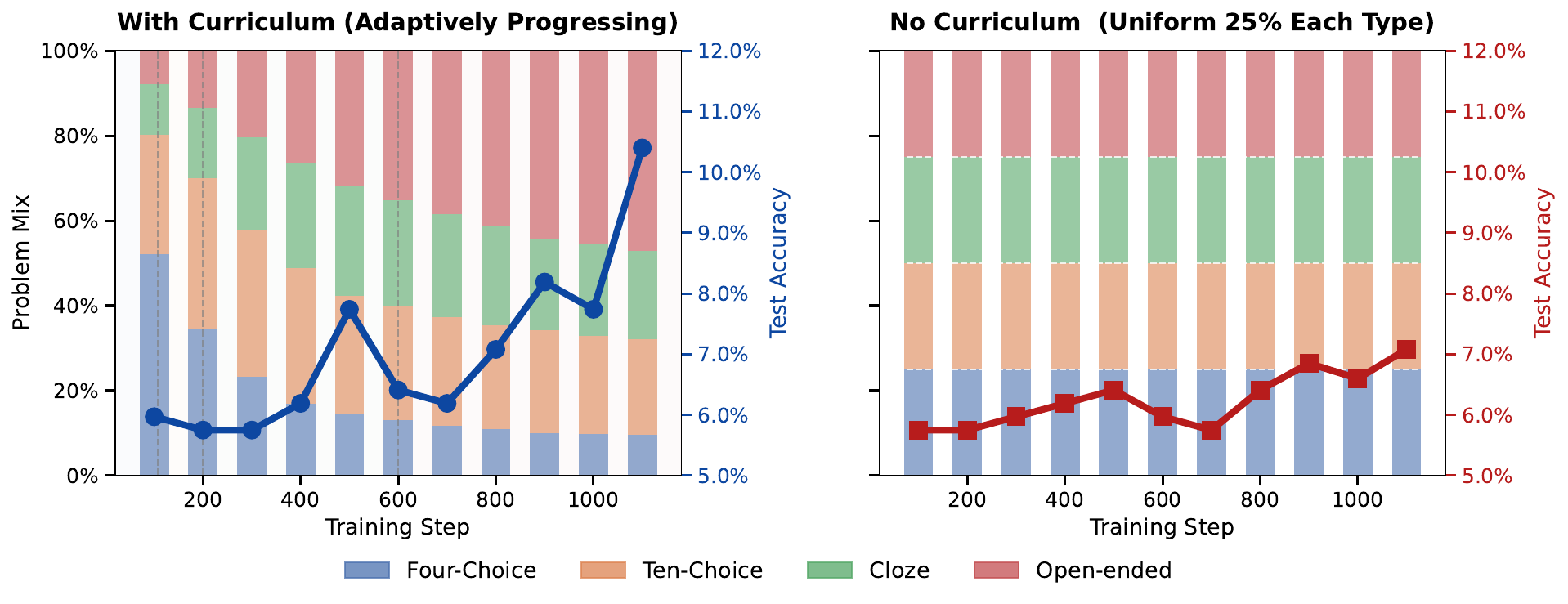}
    \caption{\textbf{Left:} Instance-level curriculum adaptively reallocates samples from easier (MCQ) to harder (OEQ) reformulations based on per-instance accuracy, leading to improved sample efficiency and continued performance gains. \textbf{Right:} A static uniform mixture (always 25\% for each format) shows stagnated performance improvement. Test accuracy is reported on open-ended questions from OmniMATH-Hard.}
    \label{fig:curriculum}
\end{figure*}

\noindent\textbf{Instance-Level curriculum improves sample efficiency.}
In \cref{tab:data_mixture}, we show that training on a mixture of multiple reformulations is already effective for learning from hard problems. However, each reformulation increases the dataset size linearly; for example, combining OEQ, 4-choice MCQ, and Cloze triples the number of training samples. Since these reformulations largely share the same underlying information (differing primarily in structure and cognitive load), we hypothesize that a more adaptive curriculum can improve sample efficiency.
In \cref{fig:curriculum}, we present both the proportion of each problem type during training and the corresponding test performance on OmniMATH-Hard, comparing settings with and without our instance-level curriculum. We refer to this approach as instance-level because progression is determined per instance: once an instance achieves an average accuracy above 50\%, it is automatically advanced to the next difficulty level. As a result, the proportion of 4-choice MCQ (the easiest format) steadily decreases over training, while intermediate formats (10-choice MCQ and Cloze) first increase and then decline, and open-ended questions grow steadily as the final stage of the curriculum.
In contrast, without the curriculum, the proportion of each problem type remains uniform throughout training (25\% each). Importantly, the curriculum leads to continued improvements in test performance on OmniMATH-Hard, whereas performance without the curriculum tends to plateau. 

\noindent\textbf{Estimating problem difficulty with pass@k can introduce selection bias.}
While using pass@k is a common way to estimate problem difficulty \citep{yuan2025llms,jiang2026fateformalbenchmarkseries,yu2025dapo}, it also comes naturally with selection bias, which includes problems that are ``truly unsolvable'' not just because the model is not capable of. 
In particular, we observe that our initial dataset $\mathcal{D}_{\text{hard}}$ contains a substantial number of flawed samples, including incomplete or unanswerable questions, as well as instances with incorrect ground-truth labels. To quantify the impact of data quality, we evaluate two filtering strategies:
(1) \textit{Binary Filter}, where GPT-5.4 \citep{gpt54} is prompted to determine whether a question is complete and answerable, and only samples labeled as solvable are retained -- without requiring the model to produce a solution; and
(2) \textit{Silver Answer Filter}, where GPT-5.4 generates three independent solutions and we retain only those samples whose ground-truth answer agrees with the majority vote.
The resulting dataset statistics and corresponding performance are summarized in \cref{tab:data_filtering}. The Silver Answer Filter is highly selective, removing a large fraction of the data, whereas the Binary Filter yields a more moderate reduction. Despite these differences, Qwen shows to be less sensitive to data quality: its performance on MATH500 remains largely unchanged across filtering strategies, with only minor variation on GPQA. We suspect that this might be relevant to its unique characteristics that can even improve with random reward \citep{shao2025spurious}. In contrast, Llama is more sensitive to dataset quality, showing a wider performance range across settings and a consistent trend of improved results with cleaner data. Therefore, we apply the Silver Answer Filter by default. Notably, GPT-5.4 is used solely to approximate the filtering of noisy question–answer pairs; we do not use its reasoning traces for any form of distillation.

\begin{table}[h]
\centering
\resizebox{\textwidth}{!}{%
\begin{tabular}{lccccc}
\toprule
 & \# Data & Qwen-MATH & Qwen-GPQA & Llama-MATH & Llama-GPQA \\
\midrule
Original $D_{\text{hard}}$ & 8922 & 82.2 & 33.1 & 24.2 & 20.7 \\
Binary Filter & 5922 & 81.8 & 34.7 & 28.0 & 22.8 \\
Silver Ans Filter & 958 & 81.6 & 36.0 & 32.2 & 23.3 \\
\bottomrule
\end{tabular}}
\caption{Impact of data quality filtering on model performance. While the Silver Answer Filter is more selective and substantially reduces dataset size, it consistently improves performance for Llama and slightly for Qwen on GPQA. Overall, Llama is more sensitive to data quality, whereas Qwen is less affected by variations in data quality.}
\label{tab:data_filtering}
\end{table}

\section{Conclusion}
In this work, we investigate how to enable language models to learn effectively from problems that lie beyond their current capabilities. Drawing inspiration from theories of human learning, we leverage cognitive load as a principled mechanism for adjusting problem difficulty, thereby bringing challenging tasks within a model's Zone of Proximal Development.
To this end, we propose \method{}, a framework that reformulates hard, open-ended questions into cognitively simplified variants that preserve the same correct answer and knowledge tested, while making them more tractable. Across two models and six challenging benchmarks, we show that \method{} not only improves pass@1 performance over strong baselines, but also generalizes effectively, and with gains extending to higher values of $k$. This suggests that the model is genuinely learning from problems that naive approaches fail to exploit.
Furthermore, we demonstrate that curriculum design plays a critical role in sample efficiency: our instance-level curriculum continues to yield improvements over time, whereas static data mixtures quickly plateau.
Overall, our results highlight problem reformulation as a simple yet powerful strategy for reducing task difficulty, offering a practical pathway to unlock learning from previously intractable problems.

\section*{Acknowledgments}
This work was supported by NSF-CAREER Award 1846185, NSF AI Engage Institute DRL-2112635, Capital One Research Award, Apple PhD Fellowship, and NDSEG PhD Fellowship. The views contained in this article are those of the authors
and not of the funding agency.

\bibliography{colm2026_conference}
\bibliographystyle{colm2026_conference}

\appendix
\input{appendix}

\end{document}

%% file: appendix.tex
\newpage
\crefalias{section}{appendix}
\section*{Appendix}

\section{Implementation Details}
\label{sec:implementation}
All methods are run on 4 A6000 GPUs. During training, we use a temperature of 1.0 and set the clip-high parameter $\epsilon$ to 0.28 \citep{yu2025dapo}. The rollout number is 8 for every GRPO-based method. We cap the output length at 8192 tokens for both training and testing. At inference, we fix the temperature to 0.7. Token limits and inference temperature are aligned across all baselines for evaluation. We report the detailed hyperparameters in \cref{tab:config}.

\begin{table}[h]
\centering
\begin{tabular}{lcc}
\toprule
 & \textbf{Llama} & \textbf{Qwen} \\
\midrule
num\_train\_samples   & 958  & 958  \\
max\_prompt\_length   & 1024  & 1024 \\
max\_response\_length & 8192  & 8192 \\
lr                    & 1e-6  & 1e-6  \\
clip\_ratio\_low      & 0.2   & 0.2   \\
clip\_ratio\_high     & 0.28  & 0.28  \\
rollout\_temperature  & 1     & 1    \\
rollout\_n            & 8    & 8   \\
kl\_coef              & 0.01 & 0.01 \\
train\_batch\_size    & 8  & 8  \\
\bottomrule
\end{tabular}
\caption{Configurations and hyperparameters we use for training.}
\label{tab:config}
\end{table}

\section{Dataset Statistics and Licenses}
\label{sec:dataset}
\method{} is tested across a diverse set of reasoning-intensive tasks spanning STEM and other domain-specific areas, including:
\textbf{(1) BigMathHard (BMH)}, a pass@64=0 test split with 192 samples from BigMath \citep{albalak2025big}.
\textbf{(2) OmniMATH-Hard}, the hardest (difficulty $\geq$ 7.5) 452 Olympiad-level math problems from OmniMATH \citep{gao2024omni};
\textbf{(3) AIME 24/25}, 30 mathematics problems from the 2024/2025 AIME competition \citep{AIME};
\textbf{(4) GPQA Diamond}, 198 PhD-level questions covering biology, physics, and chemistry \citep{rein2024gpqa};
\textbf{(5) Date Understanding}, 369 date and time arithmetic tasks from Big-Bench \citep{srivastava2022beyond}.
Some of the analyses are also run on \textbf{MATH500}, a subset 500 problems from the MATH benchmark \citep{hendrycksmath2021} curated by \citet{lightman2023let};
Note that GPQA and Date Understanding consist of multiple choice questions, but we evaluate the model with answer match \citep{chandak2025answer}, i.e., the generated answer should match the ground truth answer, instead of just comparing the option choice. This mitigates the tendency to generate a shortcut answer or memorize the option choice. For Date Understanding which questions are easier, we remove option choices to make it fully open-ended. We provide the sample sizes and licenses of the datasets used in this work in \cref{tab:datasets}. All the datasets are in English and all datasets are used in a fashion consistent with their intended use.
\begin{table*}[ht!]
    \centering
    \begin{tabular}{lcc}
    \toprule
    & Sample Size & License \\\midrule
       BigMath (Train) \citep{albalak2025big} & 958 & Apache License \\
       BigMath (Test) \citep{albalak2025big} & 192 & Apache License \\
       OmniMATH \citep{gao2024omni} & 452 & Apache License \\
       AIME 2024 \citep{AIME} & 30 & CC0 \\
       AIME 2025 \citep{AIME} & 30 & CC0 \\
       MATH500 \citep{lightman2023let,hendrycksmath2021} & 500 & MIT License \\
       GPQA-Diamond \citep{rein2024gpqa} & 198 & MIT License \\
       Date Understanding \citep{suzgun2022challenging} & 369 & Apache License\\\bottomrule
    \end{tabular}
    \caption{The statistics and licenses of the datasets used in this study. Unless otherwise specified, like BigMath has a train split, all the datasets are used for evaluation.}
    \label{tab:datasets}
\end{table*}

\newpage
\section{Prompts for Reformulation}
\label{sec:prompt}
We use the following prompts to conduct problem reformulation:

\begin{example*}[Prompt for Open-ended $\rightarrow$ Four Choice MCQ]
Please create a multiple-choice question with 4 options (A, B, C, D) based on the following math problem and its correct answer.

Original Question: \texttt{\{question\}}\\
Correct Answer: \texttt{\{gold\_answer\}}

\textbf{Requirements:}\\
1. Keep the original question\\
2. Add 4 options (A, B, C, D)\\
3. One option should be the correct answer\\
4. The other 3 options should be plausible but incorrect distractors\\
5. Randomly place the correct answer among the options\\

\textbf{Output format:}\\
Question: [the question]\\
A. [option A]\\
B. [option B]\\
C. [option C]\\
D. [option D]\\

Correct Answer: [letter of correct option]\\
Only output in this exact format, nothing else.
\end{example*}

\newpage
\begin{example*}[Prompt for Open-ended $\rightarrow$ Ten Choice MCQ]
Please create a multiple-choice question with 10 options (A, B, C, D, E, F, G, H, I, J) based on the following math problem and its correct answer.

Original Question: \texttt{\{question\}}\\
Correct Answer: \texttt{\{gold\_answer\}}

\textbf{Requirements:}\\
1. Keep the original question\\
2. Add 10 options (A, B, C, D, E, F, G, H, I, J)\\
3. One option should be the correct answer\\
4. The other 9 options should be plausible but incorrect distractors\\
5. Randomly place the correct answer among the options\\

\textbf{Output format:}\\
Question: [the question]\\
A. [option A]\\
B. [option B]\\
C. [option C]\\
D. [option D]\\
E. [option E]\\
F. [option F]\\
G. [option G]\\
H. [option H]\\
I. [option I]\\
J. [option J]\\

Correct Answer: [letter of correct option]\\
Only output in this exact format, nothing else.
\end{example*}

\begin{example*}[Prompt for Open-ended $\rightarrow$ Cloze]
Your task is to produce a masked version of the correct answer by replacing some digits with underscores (\_).

\textbf{Requirements:}\\
1. Masked answer is the correct answer with some digits replaced by underscores (\_)\\
2. Preserve LaTeX formatting in the masked answer (e.g., if answer is \texttt{\textbackslash frac\{1\}\{2\}}, mask it as \texttt{\textbackslash frac\{1\}\{\_\}} or similar)\\
3. Mask approximately 50-80\% of the digits, keeping at least one digit visible\\
4. Only mask numbers, not letters or latex symbols

\textbf{Output format:}
\textbackslash boxed\{{[masked answer with underscores only]}\}

\textbf{Examples:}
If the answer is 1003, output: \textbackslash boxed\{1\_\_3\} or \textbackslash boxed\{\_\_03\}\\
If the answer is \textbackslash frac\{5\}\{8\}, output: \textbackslash boxed\{\textbackslash frac\{5\}\{\_\}\} or \textbackslash boxed\{\textbackslash frac\{\_\}\{8\}\}\\
\\
Only output the masked answer in \textbackslash boxed\{\}, nothing else.

Original Question: \texttt{\{question\}}\\
Correct Answer: \texttt{\{gold\_answer\}}\\
Masked Answer: 
\end{example*}

\newpage
\section{Examples of Different Problem Formats}
\label{sec:problem_type}
\begin{example}[Open-ended Problem]
\textbf{Question (x):} What is the largest number of cells in a $9 \times 9$ board that can be cut along both diagonals without the board falling into pieces? \textbf{Answer (y): 21}
\end{example}

\begin{example}[Four Choice Problem]
\textbf{Reformulated Question (x'):} What is the largest number of cells in a $9 \times 9$ board that can be cut along both diagonals without the board falling into pieces? A. 18 B. 21 C. 24 D. 15 \textbf{Answer (y): 21}
\end{example}

\begin{example}[Ten Choice Problem]
\textbf{Reformulated Question (x'):} What is the largest number of cells in a $9 \times 9$ board that can be cut along both diagonals without the board falling into pieces? A. 4 B. 6 C. 7 D. 9 E. 12 F. 15 G. 18 H. 21 I. 24 J. 27 \textbf{Answer (y): 21}
\end{example}

\begin{example}[Cloze Problem]
\textbf{Reformulated Question (x'):} What is the largest number of cells in a $9 \times 9$ board that can be cut along both diagonals without the board falling into pieces? The answer should look like: 2\_. Fill the blank by giving the full answer. \textbf{Answer (y): 21}
\end{example}

\section{Example of Truly Unsolvable Problems}
\label{sec:unsolvable}
When using pass@k to collect our $\mathcal{D}_\text{hard}$, we find that a large portion of the dataset contains truly unsolvable problems. Below we list two of them:

\begin{example}[Incorrect Gold Answer]
\textbf{Question:} The cube root of a number is equal to the square root of the number. What is the number?
\textbf{Gold Answer:} 0 (Should be 0 or 1)
\end{example}

\begin{example}[Incomplete or Ambiguous Question]
\textbf{Question:} Using five sixes, arithmetic operations, and exponentiation, form numbers from 1 to 14.
\textbf{Gold Answer:} 1-14
\end{example}

\begin{example}[Lack of Figures]
\textbf{Question:} As shown in the figure, in a rectangle with an area of 10,000 square centimeters, a large semicircle and two equal small semicircles are cut out. The remaining shaded area is $\_$ square centimeters. (Take $\pi$ as 3.14)
\textbf{Gold Answer:} 2150
\end{example}

\newpage
\section{Qualitative Examples of Baselines and \method{} Outputs}
\label{sec:model_output}

\begin{example}[Output from Qwen3-4B-Instruct-2507]
\textbf{Question:} When $4444^{4444}$ is written in decimal notation, the sum of its digits is $ A.$ Let $B$ be the sum of the digits of $A$. Find the sum of the digits of $ B$ ($A$ and $B$ are written in decimal notation.)
\textbf{Gold Answer: 7}
\textbf{Reasoning:} We are given the number $ 4444^{4444} $, and we are to compute the sum of the digits of this number, call that $ A $, then the sum of the digits of $ A $  is $ B $, and finally, we are to find the sum of the digits of $ B $. Let's go step-by-step. 
Step 1: Understand what we are dealing with $ 4444^{4444} $ is an enormous  number. We cannot compute it directly. Instead, we need to find the digit sum  of this number, then digit sum of that, then digit sum again. $ A \equiv 4444^{4444} \pmod{9} $
$ B \equiv A \pmod{9} $ Then the sum of digits of $ B $ is $ \equiv B \pmod{9} $, and since the digit sum of a number is congruent to the number mod 9, we can use that to find the final answer. Moreover, since the sum of digits of a number is at most a certain value, and we are repeatedly taking digit sums (like in the process of reducing a number to its digital root), we can use the concept of digital root.
Step 2: The digital root of a number is the iterative sum of digits until a single digit is obtained. Digital root of $ 4444^{4444} $ is $ A \to B \to \text{(sum of digits of } B) $
So the final answer is the digital root of $ 4444^{4444} $. Therefore, we just need to compute: $4444^{4444} \mod 9$ Then, if the result is 0, the digital root is 9; otherwise, it's the remainder.
Step 3: Compute $ 4444 \mod 9 $ We can compute $ 4444 \mod 9 $ by summing digits: Digits of 4444: $ 4+4+4+4 = 16 $ Then $ 1+6 = 7 $ So $ 4444 \\equiv 7 \pmod{9} $ Therefore: $4444^{4444} \equiv 7^{4444} \pmod{9}$ Now we need to compute $ 7^{4444} \mod 9 $
Step 4: Let's compute powers of 7 modulo 9: $ 7^1 = 7 \mod 9 = 7 $
$ 7^2 = 49 \mod 9 = 49 - 54 = -5 \equiv 4 \mod 9 $ 
$ 7^3 = 7 \cdot 7^2 = 7 \cdot 4 = 28 \mod 9 = 1 $ 
$ 7^4 = 7 \cdot 1 = 7 \mod 9 $ Pattern: $ 7, 4, 1, 7, 4, 1, \ldots $ 
Cycle of length 3: $ 7, 4, 1 $ So the cycle is period 3 . 
Step 5: Find $ 4444 \\mod 3 $ Sum of digits of 4444: $ 4+4+4+4 = 16 $, $ 1+6 = 7 $, $ 7 \mod 3 = 1 $ So $ 4444 \equiv 1 \pmod{3} $ Therefore, in the cycle $ (7, 4, 1) $, since exponent $ 4444 \equiv 1 \pmod{3} $, it corresponds to the first position. 
Step 6: Digital root is 7. So the digital root of $ 4444^{4444} $ is 7. This means: $ A \equiv 7 \pmod{9} $
$ B \equiv 7 \pmod{9} $. The sum of digits of $ B $ is $ \equiv 7 \pmod{9} $ 
Step 7: Estimate the number of digits of $ 4444^{4444} $ Number of digits of a number $ N $ is $ \lfloor \log_{10} N \rfloor + 1 $. First compute $ 4444 \times 3.643 $ Break it down: $ 4444 \times 3 = 13332 $ 
$ 4444 \times 0.6 = 2666.4 $ 
$ 4444 \times 0.04 = 177.76 $ 
$ 4444 \times 0.003 = 13.332 $ 
Add them: $ 13332 +2666.4 = 15998.4 $, $ +177.76 = 16176.16 $, $ +13.332 = 16189.492 $ So total log $ 16189.492$ Thus, number of digits  $ 16190 $ So $ 4444^{4444} $ has about  16,190 digits  Now, the  maximum possible digit sum  $ A $ is $ 9 \times 16190 = 145,710 $ So $ A \leq 145,710 $ Now, $ B $ is the sum of digits of $ A $. Since $ A \leq 145,710 $, which is a 6-digit number. So $ B \leq 9+9+9+9+9+9 = 54 $ So $ B \leq 54 $ Now, $ B \equiv 7 \pmod{9} $. So possible values of $ B $: 7, 16, 25, 34, 43, 52. Now, for each, compute the sum of digits: $ 7 $: sum = 7 $ 16 $: $ 1+6 = 7 $, $ 25 $: $ 2+5 = 7 $, $ 34 $: $ 3+4 = 7 $, $ 43 $: $ 4+3 = 7 $, $ 52 $: $ 5+2 = 7 $  All of them have digit sum = 7! So regardless of which one it is, the sum of the digits of $ B $ is  7 \\
Final Answer: $\boxed{7}$
\end{example}

\newpage

\begin{example}[Output from Qwen trained with GRPO]
\textbf{Question:} Let $a,b,c,d$ be real numbers such that $b-d \ge 5$ and all zeros $x_1, x_2, x_3,$ and $x_4$ of the polynomial $P(x)=x^4+ax^3+bx^2+cx+d$ are real. Find the smallest value the product $(x_1^2+1)(x_2^2+1)(x_3^2+1)(x_4^2+1)$ can take.
\textbf{Gold Answer:} 16
\textbf{Reasoning:} We are given a polynomial  $P(x) = x^4 + ax^3 + bx^2 + cx + d$  with all real roots $ x_1, x_2, x_3, x_4 $, and the condition that  $b - d \geq 5.$  We are to minimize the product  $(x_1^2 + 1)(x_2^2 + 1)(x_3^2 + 1)(x_4^2 + 1),$  given this constraint.
Step 1: Understand the expression. We are to minimize  $\prod_{i=1}^4 (x_i^2 + 1).$. We can write this in terms of symmetric functions. Note that:$(x_i^2 + 1) = (x_i - i)(x_i + i),$  so the full product becomes:$\prod_{i=1}^4 (x_i^2 + 1) = \prod_{i=1}^4 (x_i - i)(x_i + i) = \prod_{i=1}^4 (x_i - i) \cdot \prod_{i=1}^4 (x_i + i) = P(i) \cdot P(-i).$. So, $\prod (x_i^2 + 1) = P(i) \cdot P(-i) = |P(i)|^2.$. 
Key observation: The product $ (x_1^2+1)(x_2^2+1)(x_3^2+1)(x_4^2+1) = |P(i)|^2 $. So, we want to minimize $ |P(i)|^2 $ over real roots $ x_1,x_2,x_3,x_4 $ such that $ b - d \geq 5 $.
Step 2: Express $ b - d $ in terms of roots. Given polynomial with real roots $ x_1,x_2,x_3,x_4 $, we write:$P(x) = (x - x_1)(x - x_2)(x - x_3)(x - x_4).$. 
We are given $ b - d \geq 5 $. So we must minimize $ |P(i)|^2 $ subject to  $\sum_{i<j} x_i x_j - x_1x_2x_3x_4 \geq 5.$. So symmetric even roots don't satisfy the constraint. We need asymmetric roots. Try three equal roots and one different?  Or two equal, and two symmetric others? Alternatively, suppose two roots are equal, and the other two are symmetric. But instead, let's suppose three roots are equal, and the fourth is different. But let's try to assume symmetry between two pairs with different signs. Suppose the roots are $ p, q, -p, -q $. That's symmetric, but we just saw that gives negative $ b - d $. Wait, we must have $ b - d \geq 5 $, and $ b - d = \sum_{i<j} x_i x_j - \prod x_i $. So we want the sum of products of pairs minus the product of all four to be at least 5. Let's denote: $ s_1 = \sum x_i $- $ s_2 = \sum_{i<j} x_i x_j $- $ s_3 = \sum_{i<j<k} x_i x_j x_k $- $ s_4 = x_1x_2x_3x_4 $. So now, we have: Minimize $ (1 - b + d)^2 + (c - a)^2 $  Subject to $ b - d \geq 5 $. Let $ x = b - d \geq 5 $. Then $ d = b - x $. So $ 1 - b + d = 1 - b + (b - x) = 1 - x $. So $ (1 - b + d)^2 = (1 - x)^2 $. Now, the second term is minimized when $ x = 5 $. Thus, the smallest possible value of the expression is $ (5 - 1)^2 = 16 $. So the product $ (x_1^2+1)(x_2^2+1)(x_3^2+1)(x_4^2+1) = |P(i)|^2 $ can be as small as 16, when $ b - d = 5 $, and $ c = a $. 
Step 4: Construct a polynomial with real roots, satisfying $ b - d = 5 $, $ c = a $. We now try to construct such a polynomial. We want: $ b - d = 5 $- $ a = c $. We are free to choose the roots as long as they are real. Let us suppose two roots are equal, and the other two symmetric. Try roots: $ p, p, q, q $. So symmetric. Then the polynomial is: $(x - p)^2 (x - q)^2 = (x^2 - 2p x + p^2)(x^2 - 2q x + q^2)$. Assume $ p + q \geq 0 $, divide both sides:$1 = pq$. So $ pq = 1 $. Now, $ b - d = 5 $. We compute $ b - d $: $ b = p^2 + 4pq + q^2 = p^2 + q^2 + 4 $- $ d = p^2 q^2 = (pq)^2 = 1 $. So $ b - d = p^2 + q^2 + 4 - 1 = p^2 + q^2 + 3 $. Set equal to 5: $p^2 + q^2 + 3 = 5 \Rightarrow p^2 + q^2 = 2$. But $ pq = 1 $, so we have: $ p^2 + q^2 = 2 $- $ pq = 1 $. Try $ p = q = 1 $. Then roots: $ 1,1,1,1 $, all equal. Polynomial: $(x-1)^4 $. Check $ a, b, c, d $: $ a = -4 $- $ b = 6 $- $ c = -4 $- $ d = 1 $. So $ b - d = 6 - 1 = 5 $. And $ c = -4 $, $ a = -4 $, $ a = c $. Perfect! Now compute the product: $(x_1^2 + 1)(x_2^2 + 1)(x_3^2 + 1)(x_4^2 + 1) = (1^2 + 1)^4 = (2)^4 = 16$. Exactly what we wanted! And all roots are real (in fact, all equal). And $ b - d = 5 $, which satisfies the condition. Therefore, the minimal value is achievable. Final Answer: \boxed{16}
\end{example}

\newpage

\begin{example}[Output from Qwen trained with \method{}]
\textbf{Question:} Let $n$ be a positive integer. Find all $n \times n$ real matrices $A$ with only real eigenvalues satisfying $A+A^{k}=A^{T}$ for some integer $k \geq n$. \textbf{Gold Answer:} 0 \\
\textbf{Reasoning:} We are given a condition for an $ n \times n $ real matrix $ A $: $ A + A^k = A^T$ for some integer $ k \geq n $, and we are to find all such matrices with real eigenvalues.
Step 1: Understand the given equation\\
Given:. $ A + A^k = A^T \quad \text{(1)}$. Rewriting it: $ A^T = A + A^k $. So the transpose of $ A $ is equal to $ A $ plus a higher power of $ A $ \\
Analyze the equation: Since we are told that $ k \geq n $, we are dealing with a high-order power (at least the size of the matrix) The transpose is expressed in terms of the matrix and its power. \\
Step 2: Use transpose properties. From equation (1): $ A^T = A + A^k $, take the transpose of both sides: $ A = A^T + (A^k)^T = (A + A^k)^T = A^T + (A^k)^T = A^T + A^{k^T}$. But $ (A^k)^T = (A^T)^k $, so: $ A = A^T + (A^T)^k. $ But from original, $ A^T = A + A^k $, so substitute that in: $ A = (A + A^k) + (A^T)^k = A + A^k + (A^T)^k. $. Now subtract $ A $ from both sides: $ 0 = A^k + (A^T)^k. $ So we are led to the key identity: $ A^k + (A^T)^k = 0 \quad \text{(2)}$. 
Now use this key equation. Given that $ A^k + (A^T)^k = 0 $, we write: $ A^k = - (A^T)^k. $. Now observe that $ (A^T)^k = (A^k)^T $, so: $ A^k = - (A^k)^T$. So the matrix $ A^k $ is skew-symmetric!\\
Important property: $ A^k $ is skew-symmetric. But we are told that $ A $ is a real matrix with real eigenvalues. So the eigenvalues of $ A $ are real. But the matrix $ A^k $ is skew-symmetric. Skew-symmetric matrices have purely imaginary eigenvalues (in general), unless zero.\\
So: $ A^k $ is skew-symmetric. $ A $ has real eigenvalues. Now, for a real matrix with real eigenvalues, $ A^k $ will also have real eigenvalues (since real powers of real matrices are real). But skew-symmetric real matrices have purely imaginary eigenvalues. So we have: $ A^k $ is skew-symmetric eigenvalues are $ \pm i \times \text{real} $. But $ A^k $ has real eigenvalues (since A has real eigenvalues). So the only way both happen is: $ A^k = 0 $ Because zero matrix is both symmetric and skew-symmetric. So we conclude: $ A^k = 0 $ Then from original equation: $ A + 0 = A^T \Rightarrow A^T = A $So $ A $ is symmetric!\\
But we must verify if this is the only solution. So: $ A^k = 0 $ for some $ k \geq n $ - and $ A^T = A$ Therefore, we are to find symmetric real matrices with zero power for some $ k \geq n $. But symmetric matrices are diagonalizable over real numbers. So if $ A $ is symmetric and nilpotent, then: $ A^k = 0 $ for some $ k \geq 1 $. But for diagonalizable matrix, $ A^k = 0 $ implies all eigenvalues are 0. So $ A = 0 $. So is the zero matrix a solution? Check: $ A + A^k = 0 + 0 = 0 $ - $ A^T = 0 $. So yes, $ 0 = 0 $: true.\\
But are there other solutions? We derived that $ A^k $ is skew-symmetric and has real eigenvalues must be zero. So $ A^k = 0 $, and then $ A^T = A $ So the matrix is symmetric and nilpotent. 
Therefore, the only such matrix is the zero matrix. So, conclusion: The only real matrix with real eigenvalues satisfying $ A + A^k = A^T $ for some $ k \geq n $ is the zero matrix. But let's verify this.\\
Verification: Let $ A = 0 $- $ A + A^k = 0 + 0 = 0 $ - $ A^T = 0 $ So it works. Are there other matrices? Suppose $ A $ is symmetric and $ A^k = 0 $Then $ A $ is diagonalizable with eigenvalues $ 0 $ so $ A = 0 $ So only zero matrix. But what about other values of $ k $? For example, if $ A^2 = 0 $, and $ A + A^2 = A^T $, then: $ A + 0 = A^T \Rightarrow A^T = A $ So again, symmetric and nilpotent. So for any $ k \geq n $, if $ A^k $ is skew-symmetric and eigenvalues are real only zero. So the condition forces: $ A^T = A $ - $ A^k = 0 $ for some $ k \geq n $ Therefore, only the zero matrix. So final boxed answer:  $\boxed{0}$.
\end{example}